%% file: main_pami.tex
\documentclass[10pt,journal,compsoc]{IEEEtran}
\usepackage[utf8]{inputenc} 
\usepackage[T1]{fontenc}  

\usepackage[breaklinks=true,
            colorlinks,
            linkcolor = red,
            urlcolor  = purple, 
            citecolor = blue,
            bookmarks = black]{hyperref}  
\usepackage{url}  
\usepackage{amssymb}   
\usepackage{booktabs} 
\usepackage{amsfonts}  
\usepackage{nicefrac}  
\usepackage{microtype}  
\usepackage{wrapfig}
\usepackage{bbding}
\usepackage{appendix}
\usepackage{graphicx}
\usepackage{amsmath}
\usepackage{amsthm}
\usepackage{subcaption}
\usepackage{float}
\usepackage{appendix}
\usepackage{xcolor}
\usepackage{multirow}

\usepackage{bm}
\usepackage{bbm}
\usepackage{makecell}
\usepackage{mathtools}
\usepackage{algorithmic}
\usepackage{algorithm}
\usepackage[algo2e,linesnumbered]{algorithm2e}

\ifCLASSOPTIONcompsoc
\usepackage[nocompress]{cite}
\else
\usepackage{cite}
\fi

\ifCLASSINFOpdf
\usepackage{graphicx}
\usepackage{amsmath,amssymb}
\usepackage{color}
\usepackage{multirow}
\usepackage{booktabs}
\usepackage[T1]{fontenc}
\usepackage{float}
\usepackage{color}
\usepackage{setspace}
\usepackage{url}
\usepackage{makecell}
\usepackage{cleveref}
\usepackage{hyperref}
\usepackage{adjustbox}
\usepackage{multirow}
\usepackage{bbm}
\usepackage{bm}
\usepackage{threeparttable}
\usepackage{bbding}
\usepackage{booktabs}
\usepackage{color}
\usepackage{amssymb}
\usepackage{cuted}
\usepackage{capt-of}

\renewcommand{\paragraph}[1]{\vspace{.25em}\noindent\textbf{#1.}}

\begin{document}

\title{Stereo-Talker: Audio-driven 3D Human Synthesis with Prior-Guided Mixture-of-Experts}

\author{Xiang Deng*,
        Youxin Pang*,
        Xiaochen Zhao*,
        Chao Xu,
        Lizhen Wang,
        Hongjiang Xiao,
        \\
        Shi Yan,
        Hongwen Zhang,
        and Yebin ${\rm{Liu}^{\dag}}$,~\IEEEmembership{Member,~IEEE}
\IEEEcompsocitemizethanks{
\IEEEcompsocthanksitem Xiang Deng, Youxin Pang, Xiaochen zhao, Chao Xu and Yebin Liu are with the Department of Automation, Tsinghua University, Beijing 100084, China. E-mail:\{dx23, pyx24,zhaoxc19\}@mails.tsinghua.edu.cn; lyxc127@outlook.com; liuyebin@mail.tsinghua.edu.cn 
\IEEEcompsocthanksitem Lizhen Wang is with rhe Bytedance Inc., Beijing 100190, China. E-mail: wanglizhen.2024@bytedance.com
\IEEEcompsocthanksitem Hongjiang Xiao is with the State Key Laboratory of Media Convergence and Communication, Communication University of China, Beijing 100024, China. E-mail: xiaohj@cuc.edu.cn
\IEEEcompsocthanksitem Shi Yan is with Apple Inc, Cupertino, CA 95014. Email: neycyanshi@foxmail.com
\IEEEcompsocthanksitem Hongwen Zhang is with the School of Artificial Intelligence, Beijing Normal University, Beijing 100875, China. E-mail: zhanghongwen@bnu.edu.cn
\IEEEcompsocthanksitem *: Equal contribution
\IEEEcompsocthanksitem ${\dag}$: Corresponding author
}
}



\IEEEtitleabstractindextext{%
	\begin{abstract}
This paper introduces Stereo-Talker, a novel one-shot audio-driven human video synthesis system that generates 3D talking videos with precise lip synchronization, expressive body gestures, temporally consistent photo-realistic quality, and continuous viewpoint control. The process follows a two-stage approach. In the first stage, the system maps audio input to high-fidelity motion sequences, encompassing upper-body gestures and facial expressions. To enrich motion diversity and authenticity, large language model (LLM) priors are integrated with text-aligned semantic audio features, leveraging LLMs’ cross-modal generalization power to enhance motion quality. In the second stage, we improve diffusion-based video generation models by incorporating a prior-guided Mixture-of-Experts (MoE) mechanism: a view-guided MoE focuses on view-specific attributes, while a mask-guided MoE enhances region-based rendering stability. Additionally, a mask prediction module is devised to derive human masks from motion data, enhancing the stability and accuracy of masks and enabling mask guiding during inference. We also introduce a comprehensive human video dataset with 2,203 identities, covering diverse body gestures and detailed annotations, facilitating broad generalization. The code, data, and pre-trained models will be released for research purposes.
	\end{abstract}
	}

\maketitle

\IEEEdisplaynontitleabstractindextext

%
\IEEEpeerreviewmaketitle
\input{sections/1_intro}
\input{sections/2_related}
\input{sections/3_method}

\input{sections/4_experiments}

\input{sections/5_conclusion}





\bibliographystyle{IEEEtran}
\bibliography{IEEEbib}
\end{document}

%% file: sections/1_intro.tex
\section{Introduction}
The automatic synthesis of lifelike videos depicting a 3D speaking individual, driven by audio input and a single reference image, presents a groundbreaking advancement with profound implications across a myriad of domains, including film-making, augmenting human-computer interaction paradigms, and virtual reality.
Intuitively, the pivotal aspects of audio-driven human synthesis comprise: 1) generating high-fidelity bodily movements and facial expressions, including precise lip synchronization, vivid emotions, and a broad range of gestures; 2) yielding photo-realistic and temporally coherent talking videos; and 3) supporting the continuous view-point control while keeping 3D consistency.

Early attempts towards addressing audio-driven human synthesis often concentrate on the generation of mouth region coherent with guiding audio signals, neglecting other body parts \cite{prajwal2020lip, song2022talking, sun2022masked, cheng2022videoretalking}. Benefiting from the progress of face reenactment methods \cite{bounareli2023hyperreenact, hsu2022dual}, producing full-head talking videos derived from single portrait images has been widely explored \cite{yu2023talking, zhang2023sadtalker, shen2023difftalk}.
Recently, Corona et al. \cite{corona2024vlogger} present an approach that leverages a single portrait image and an audio clip to synthesize talking human videos. This is achieved by firstly mapping audio input to corresponding body gestures and facial expressions, followed by rendering these predicted motions utilizing advanced human video generation methodologies \cite{hu2023animate, zhang2023adding}. 
Although achieving remarkable audio-conditional image-to-video synthesis, it frequently generates unrealistic or blurry visual artifacts in the face and hand regions and does not explore view controllability.

\begin{figure*}
    \centering
    \includegraphics[width = 1\textwidth]{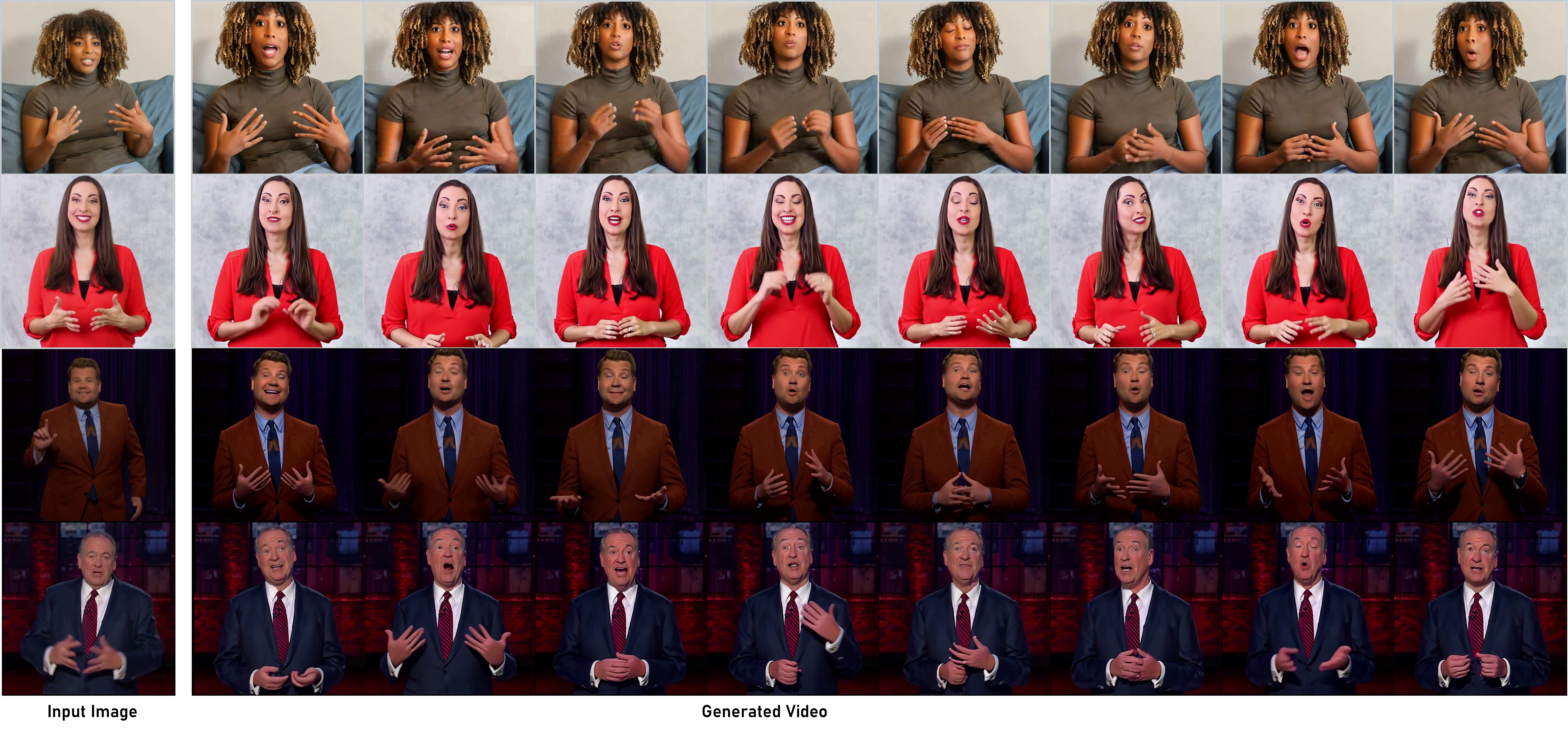}
    \caption{We present a framework designed for the synthesis of human videos driven by audio inputs. Given a single reference image in the first column and an arbitrary audio clip, our methodology produces high-fidelity photo-realistic video outputs depicting the subject engaged in realistic conversation. 
  The synthesized frames illustrate our achievement of accurate lip synchronization, spontaneous eye blinking, and vivid body gestures, collectively pushing the boundaries of audio-driven human synthesis to new heights.}
    \label{fig:teaser}
\end{figure*}

We present Stereo-Talker, a novel framework that aims for audio-driven, view-controllable human video synthesis. To extend the generation capabilities of Diffusion-based models in human body novel-view synthesis, we incorporate view-guided Mixture-of-Experts (MoE) modules\cite{shazeer2017outrageously, fedus2022switch, xue2023raphael}, to inject camera-related prior information into the generation network. Each expert within the MoE is assigned a specific viewpoint and is responsible for depicting the unique view-related human appearance. The diffusion rendering network effectively fuses the appearance information embedded in each expert, based on the distance between the given viewpoint and each expert's viewpoint. The view-guided MoE equips the generator with the capacity to learn human appearance prior knowledge under various camera perspectives, without incurring significant computational costs during training and inference. Our experiments demonstrate that view-guided MoE plays a key role in maintaining consistency across viewpoints.
Additionally, we introduce mask-guided MoE to enhance the generation of human limb details. The mask-guided MoE is responsible for depicting different body parts in human images, where each expert is guided by a segmented region mask. This improves the rendering quality by facilitating the model to distinguish different body parts effectively.
We further integrate a Variational Autoencoder (VAE) \cite{kingma2014auto} network to dynamically generate human masks at inference time, which enables accurate mask guidance at inference time and ultimately contributes to increased realism and stability in the output videos.

Moreover, most audio-driven human video synthesis approaches posit that body movements are mainly correlated to low-level rhythmic audio features \cite{corona2024vlogger, liu2022audio}. In contrast, recent studies \cite{ao2022rhythmic, liu2022disco} have emphasized the critical role of high-level semantic content in enriching and diversifying body gesture generation. Consequently, this oversimplified hypothesis leads to a restricted variety in the produced motion sequences. To address this issue, 
we propose integrating powerful large language model (LLM) priors within our audio-driven gesture synthesis framework. Aligning speech features with linguistic representations bolsters our comprehension of the contextual nuances inherent within the spoken content.
Specifically, our methodology first employs a pre-trained audio feature extractor to distill high-dimensional semantic information from the input speech, subsequently projecting these features into the latent space of large language models. 
Subsequently, these language-aligned features are utilized by large language models to augment semantic understanding. 
Finally, the enriched semantic features serve as conditional inputs to diffusion models, guiding the generation of gestures. The incorporation of large language model priors empowers our system with the capability to synthesize gestures that are not only coherent with the paired audio but also exhibit an elevated degree of stability and diversity.

To ensure the generalization ability of our framework, we also present a large-scale High-definition Audio-Visual dataset (HDAV) containing 2,203 identities with 3D human template parameter annotations and detail property labels.

In summary, our major contributions are as follows:
\begin{itemize}
    \item We present the first framework that supports the one-shot generation of high-fidelity 3D talking human videos from audio inputs.
    \item We devise a prior-guided MoE module, which improves the overall visual rendering quality and stability without largely increasing computational cost.
    \item We leverage the powerful large language model priors to guide cross-modal translation and elevate the diversity of gesture generation.
    \item We release a large-scale dataset that significantly lowers the hurdles associated with training robust models for human video generation.
\end{itemize}

%% file: sections/2_related.tex
\begin{figure*}
    \centering
    \includegraphics[width = 1\textwidth]{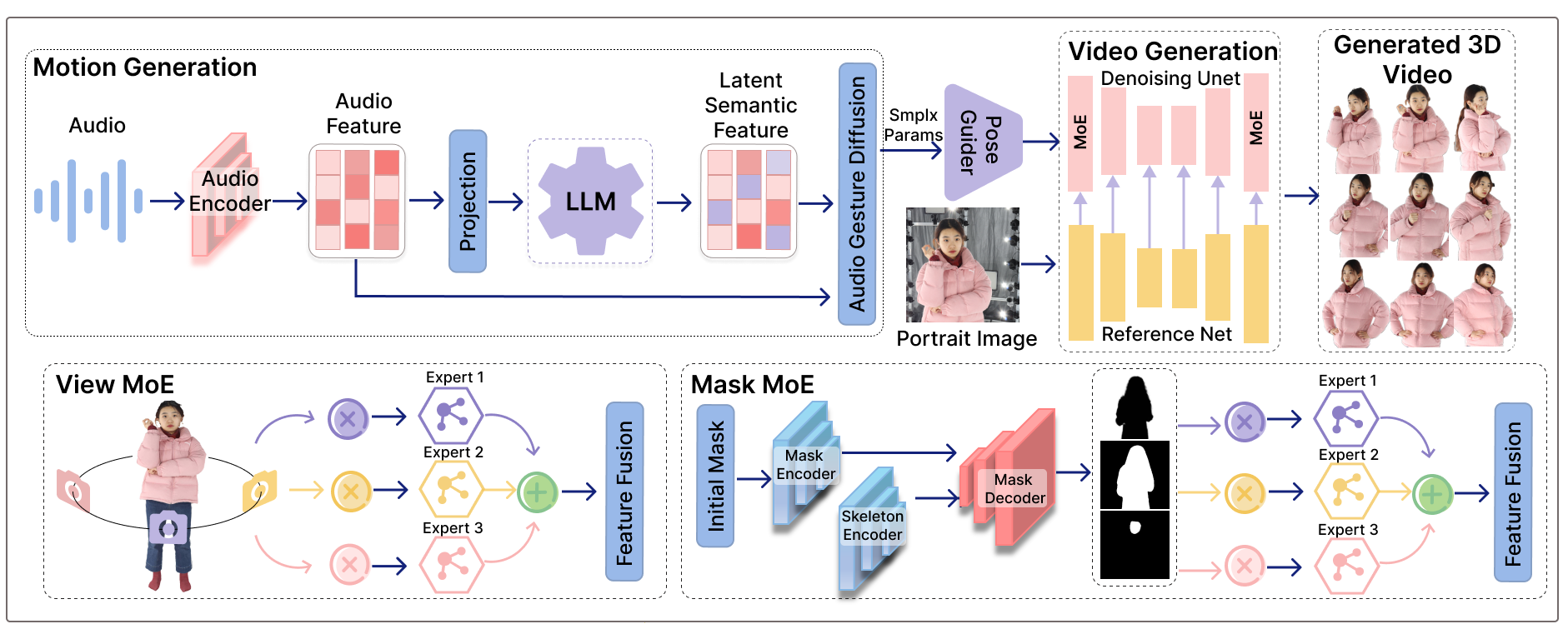}
    \caption{The overall framework of Stereo-Talker. Given a single portrait image with its driven audio, we first convert the audio input to human motion sequences based on large language model priors. Then, we render these motions to high-fidelity human videos through a U-net backbone, where a view Mixture-of-Experts (MoE) module and a mask MoE module improve the rendering stability. Notably, we train a mask generation network to predict the human mask at inference time.}
    \label{fig:main}
\end{figure*}

\section{RELATED WORK}
\subsection{Video Generation.}
The synthesis of video content has been approached using a variety of frameworks, including autoregressive transformers \cite{weissenborn2019scaling, wu2022nuwa}, recurrent neural networks \cite{babaeizadeh2018stochastic, franceschi2020stochastic, castrejon2019improved}, normalizing flows \cite{blattmann2021ipoke, dorkenwald2021stochastic}, and generative adversarial networks \cite{villegas2017decomposing, saito2020train, skorokhodov2022stylegan}. 
Benefiting from the robust and scalable training objective, diffusion models have shown their advantages in video generation \cite{blattmann2023align, blattmann2023stable, zhang2023adding, ji2022eamm}.
Voleti \textit{et. al.}  \cite{voleti2022mcvd} devise a general framework that simultaneously enables video generation, prediction, and interpolation using a probabilistic conditional score-based denoising diffusion model.
Harvey \textit{et. al.} \cite{harvey2022flexible} introduce a generative model that can produce long-duration video by generating sparse frames first, followed by the interpolation of the missing frames to ensure a smooth and continuous visual narrative.
Hu \textit{et. al.} \cite{hu2023animate} propose to merge detailed features via spatial attention and generate high-quality human videos with human skeleton sequences as conditional input.
Zhu \textit{et. al.} \cite{zhu2024champ} further explore to improve shape alignment and pose guidance mechanisms by leveraging a 3D human parametric model to enhance shape alignment and motion guidance in diffusion frameworks.
Zhang \textit{et. al.} \cite{zhang2024mimicmotion} introduce confidence aware pose guidance and regional loss amplification to improve high frame quality and temporal smoothness.
Shao \textit{et. al.} \cite{shao2024human4dit} introduce diffusion transformers to human video generation and propose a 4D diffusion transformer architecture to model 
 both spatial and temporal correlations between body parts.
Corona \textit{et. al.} \cite{corona2024vlogger} make the first attempt to bridge the gap between video diffusion models and talking-human generation.
However, these methods typically render the whole image uniformly, while ignoring the specific properties of different views and image regions.
Inspired by the success of Mixture of Experts (MoE) \cite{shazeer2017outrageously, fedus2022switch, xue2023raphael},
we incorporate a prior-guided MoE module within the diffusion framework, thereby improving the differentiation among diverse image regions. This strategic integration enhances not only the quality of generated outputs but also improves the stability of the generation process.


\subsection{Co-speech Gesture Generation.}
Pioneering efforts in co-speech gesture generation typically rely on linguistic rules that map audio data to pre-defined gesture segments and generate gestures in a search-and-connect manner \cite{kipp2005gesture, kopp2006towards, cassell1994animated}.
However, rule-based methods entail substantial endeavors in collecting gesture data and designing intricate rules. To address this limitation, researchers explore learning the mapping relationship between audio and gestures using deep neural networks \cite{kucherenko2020gesticulator, yoon2019robots, ahuja2019language2pose, danvevcek2023emotional}.
With the aim of enhancing expressiveness in generated outputs, researchers propose to predict facial landmarks in addition to body gestures \cite{liu2022beat, liu2023emage}, thus enriching the realism and emotional nuance in the synthesized content.
These methods typically model co-gesture generation in a deterministic manner \cite{liu2022learning, pang2023bodyformer, qi2023weakly} and ignore the many-to-many relationship between speech and gesture, leading to challenges in achieving a precise alignment of meaning between the generated gestures and the driven audio \cite{yoon2022genea}.
In response to this challenge, methods based on probabilistic models have been widely explored \cite{ginosar2019learning, yoon2020speech, zhu2023taming, ao2023gesturediffuclip, li2021audio2gestures, qian2021speech, sun2023co, yang2023qpgesture, alexanderson2023listen}.
In pursuit of high-quality visual realism, researchers \cite{liu2022audio, he2024co, liu2023moda} further propose unified frameworks designed to synthesize co-speech gesture videos under audio guidance. 
Guo \cite{guo2024fasttalker} further explore to simultaneously generates high-quality speech audio and 3D human gestures from text.
However, most methods simply hypothesize that human skeletons are mainly correlated with rhythm and beat and generate human skeletons based on only low-dimensional rhythm features, which limits their expressive capability.
To fill this gap, researchers explore involving both rhythmic and semantic features to enhance the coherence between generated gestures and the associated speech content \cite{ao2022rhythmic, liu2022disco}.
Nevertheless, these methodologies learn distinctions between low-level rhythmic elements and high-level semantic features within a narrow dataset scope, consequently yielding sub-optimal results. 
To enhance generalization, Zhang \textit{et al.} \cite{Zhang2024SemanticGesture} leverage large language models (LLM) to retrieve semantic gestures from a pre-established motion library. Nonetheless, the motion library was constructed using a dataset with a limited number of identities, which may constrain its robustness when encountering unseen individuals.
Capitalizing on the recent progress in employing LLM priors to facilitate cross-modal alignment tasks \cite{wu2023nextgpt, jiang2023motiongpt}, we view co-gesture generation as a language translation problem and borrow the power of LLM priors to significantly enhance the fidelity and diversity of gesture synthesis.

%% file: sections/3_method.tex
\section{Method}
This section presents Stereo-Talker, a novel method that enables the 3D generation of 
a speaking individual depending on the provided audio and a single reference image.
The pipeline of the overall framework is depicted in Figure \ref{fig:main}.
We first introduce how to employ pre-trained large language models to establish a sophisticated mapping between raw audio and human motion sequences in Section \ref{Method_audio}. Then, we introduce the design of our proposed view-guided and mask-guided Mixture of Experts (MoE) module in Section \ref{Method_moe}, which aims at augmenting both the authenticity and stability of the synthesized video output, thereby pushing the boundaries of video generation quality.
After that, we describe the process of rendering the generated human motion into visually authentic speaking videos in Section \ref{Method_render}. 
Finally, we describe the detail property of the proposed large-scale human video dataset in Section \ref{Method_dataset}.

Given a single portrait image and a sequential audio input \(\textbf{a} = [a_1, ..., a_N]\), our method aims to synthesize a coherent talking video sequence  \(\textbf{x}\) = \([x_1, ..., x_N]\), which not only ensures synchronization of lip motions with the speech but also incorporates body gestures that are harmoniously aligned with the nuances of the driving audio, thereby achieving a highly immersive and naturalistic audio-visual experience.
As shown in Figure \ref{fig:main}, our approach accomplishes this in a two-stage process. Firstly, we leverage powerful large language models to convert the audio sequence \(\textbf{a}\) into a series of per-frame human pose representations, denoted as \(\textbf{p} = [p_1, ..., p_N]\). Following this, we employ the video diffusion technique to render these pose frames \(\textbf{p}\) into photo-realistic talking video frames \(\textbf{x}\).


\subsection{LLM-enhanced Audio-Driven Motion Generation}
\label{Method_audio}
Most prior studies \cite{corona2024vlogger, zhu2023taming} posit that body gestures mainly correlate with speech rhythm and can be adequately produced using low-dimensional rhythm features. 
Conversely, drawing inspiration from the remarkable success of large language models (LLMs) in bridging different modalities \cite{wu2023nextgpt, jiang2023motiongpt}, we view co-speech motion generation problem as a language translation problem and leverages the large language model priors to elevate the diversity of our generated motion sequences in a similar way. 
Specifically, our approach firstly extracts high-level semantic features of audio and corresponding motion sequences, then aligns these features with the textual feature domain, where the semantic latent space is comprehensible to the core LLM. 
Subsequently, we employ a diffusion model to achieve cross-modal semantic mapping, leveraging the capabilities of the LLM to ensure coherent integration across modalities.

\subsubsection{Audio Representation Learning}
To capture high-level semantic features from audio, we employ the pre-trained wav2vec 2.0 speech model \cite{baevski2020wav2vec}. This model comprises two core components: an audio feature extractor and a transformer encoder. The feature extractor utilizes a series of temporal convolutional layers to process raw audio input \(\textbf{a}\), transforming it into a sequence of feature vectors. These feature vectors are then passed to a transformer encoder, composed of stacked multi-head attention layers, which enriches the features with contextual information, producing a sequence of contextualized latent speech representations.

\subsubsection{Motion Representation Learning}
To acquire high-fidelity and diversity motion representations, we adopt a VQ-VAE \cite{van2017neural} model, as demonstrated in previous studies \cite{yi2023generating,ao2023gesturediffuclip}, to reduce redundancy in raw motion representations. Specifically, the encoder $\mathcal{E}_p$ processes the motion sequence \(\textbf{p}\) and transforms it into a downsampled latent sequence \(\textbf{z}_p\). Each latent vector in \(\textbf{z}_p\) is then quantized by mapping it to the nearest vector in a finite codebook. The decoder $\mathcal{D}_p$ is trained to reconstruct the original human motion sequence from this quantized representation. By doing so, we ensure the preservation of essential motion information while filtering out irrelevant details.

\subsubsection{Cross-Modal Representation Mapping}
After obtaining high-quality audio and motion representations, we address the translation from verbal to non-verbal bodily language within a diffusion-conditional framework, leveraging large language models (LLMs) as a foundational prior. Since the high-level audio semantic features cannot be directly processed by LLMs due to domain discrepancies, we introduce a projection network that acts as a semantic translator. This network transforms the rich semantic content from the high-level speech features into a representation compatible with the textual latent space of the LLM. Given the semantic richness of the audio features, a lightweight projection network suffices to align the features with the LLM’s latent space. This streamlined approach ensures semantic congruence while maintaining computational efficiency.
To further enhance semantic correlation and improve feature quality, we utilize a pre-trained LLM encoder \cite{raffel2020exploring, chung2024scaling} as the language model backbone, fine-tuning it for co-speech gesture generation using LoRA \cite{hu2022lora}. This process is expressed as:
\begin{equation}
    \textbf{z}= \mathcal{E}_t(\mathcal{M}(\mathcal{E}_a(\textbf{a}) ), \mathcal{E}_a(\textbf{a}))
\end{equation}
where \(\mathcal{E}_t\) denotes the pre-trained LLM encoder, \(\mathcal{M}\) the projection network, \(\mathcal{E}_a\) the pre-trained audio encoder, and \(\textbf{z} = [z_1, ..., z_N]\) represents the enriched latent features. 
Notably, our method incorporates only the encoder component of the LLM, leveraging this powerful pre-trained model to refine feature representations. A diffusion network is then employed to decode these refined features into motion representations, as generative models are better suited for one-to-many mappings, thus improving both the fidelity and diversity of the generated motion sequences.
Specifically, we employ a denoising diffusion probabilistic model \cite{ho2020denoising, zhu2023taming, corona2024vlogger}, which incrementally adds Gaussian noise \(\bm{\epsilon}\) to the ground-truth samples \(\textbf{p}_0\), as defined by:
\begin{equation}
    \mathcal{L}_p = \mathbbm{E}_{\epsilon, \textbf{p}_t, \textbf{c}, t} \left( || \bm{\epsilon} - \bm{\epsilon}_{\theta}(\textbf{p}_{t}, \textbf{c}, t)|| \right)
\end{equation}
where \(\textbf{p}_{t}\) represents the noisy pose sequences at timestep \textit{t}, and \(\textbf{c}\) includes  the initial pose \(p_1\), the latent feature \(\textbf{z}\), and original audio feature $\mathcal{E}_a(\textbf{a})$ 
 to reverse audio modality unique features. By doing so, we harness the capabilities of pre-trained large language models to generate richer and more sophisticated gestures, thereby enhancing the overall expressiveness and naturalness of the synthesized motions.

\subsection{View-guided MoE and Mask-guided MoE}
\label{Method_moe}
Generating high-fidelity photo-realistic human videos based on driven gestures is an important yet challenging task.
Capitalizing on the recent breakthroughs in video diffusion models \cite{blattmann2023align, zhang2023adding}, methodologies for human video generation grounded in stable diffusion processes have garnered extensive exploration \cite{hu2023animate, xu2023magicanimate}. 
Nonetheless, these systems render the complete image uniformly depending on the U-net backbone, inadvertently disregarding the unique characteristics inherent in distinct viewpoints and body regions.
To tackle this challenge, we introduce a view-guided and mask-guided Mixture-of-Experts (MoE) mechanism designed to accommodate the distinctive traits of varying viewpoints and image segments. 

\subsubsection{View-guided MoE}
To inject view prior information into the diffusion backbone, we design a
view-guided Mixture-of-Experts module to add multiple experts, each specializing in a specific viewpoint.
For a given view, we first compute a distance matrix that quantifies the similarity between the input view and the corresponding expert. Next, a multi-layer perceptron (MLP) is employed to extract view embeddings from the distance matrix. These view embeddings are then integrated into each expert through a cross-attention layer, which injects the view-specific information into the expert’s processing pipeline. This approach ensures that each expert adapts its generative process to the unique characteristics of the given viewpoint, thereby enhancing the accuracy and coherence of the synthesized outputs.
\subsubsection{Mask-guided MoE}
To accommodate the distinctive traits of varying image segments, we segment the entire image into three distinct parts termed as face, body, and background by leveraging the facial landmarks detected by DWPose \cite{yang2023effective} and human body masks detected by RVM \cite{lin2022robust}. 
Subsequently, we integrate three separate sparse expert layers, each acquiring the corresponding region mask as a latent feature mask and acting as a specialist catering to the unique demands of its respective image portion.
Considering the instability issues commonly encountered with detected masks and their unavailability during inference,
we devise an auxiliary lightweight module trained to estimate human masks directly from skeletal data. This module comprises a dual-encoder and a single decoder. The dual-encoder setup is intended to process human mask images and human skeleton images separately, while the decoder consolidates this encoded information to reconstruct human masks. By generating high-quality masks and incorporating this information into the diffusion framework through MoE, the rendering model is able to distinguish the background, body, and face parts, thus improving the temporal consistency and stability of output videos.
\begin{figure*}
    \centering
    \includegraphics[width = 1\textwidth]{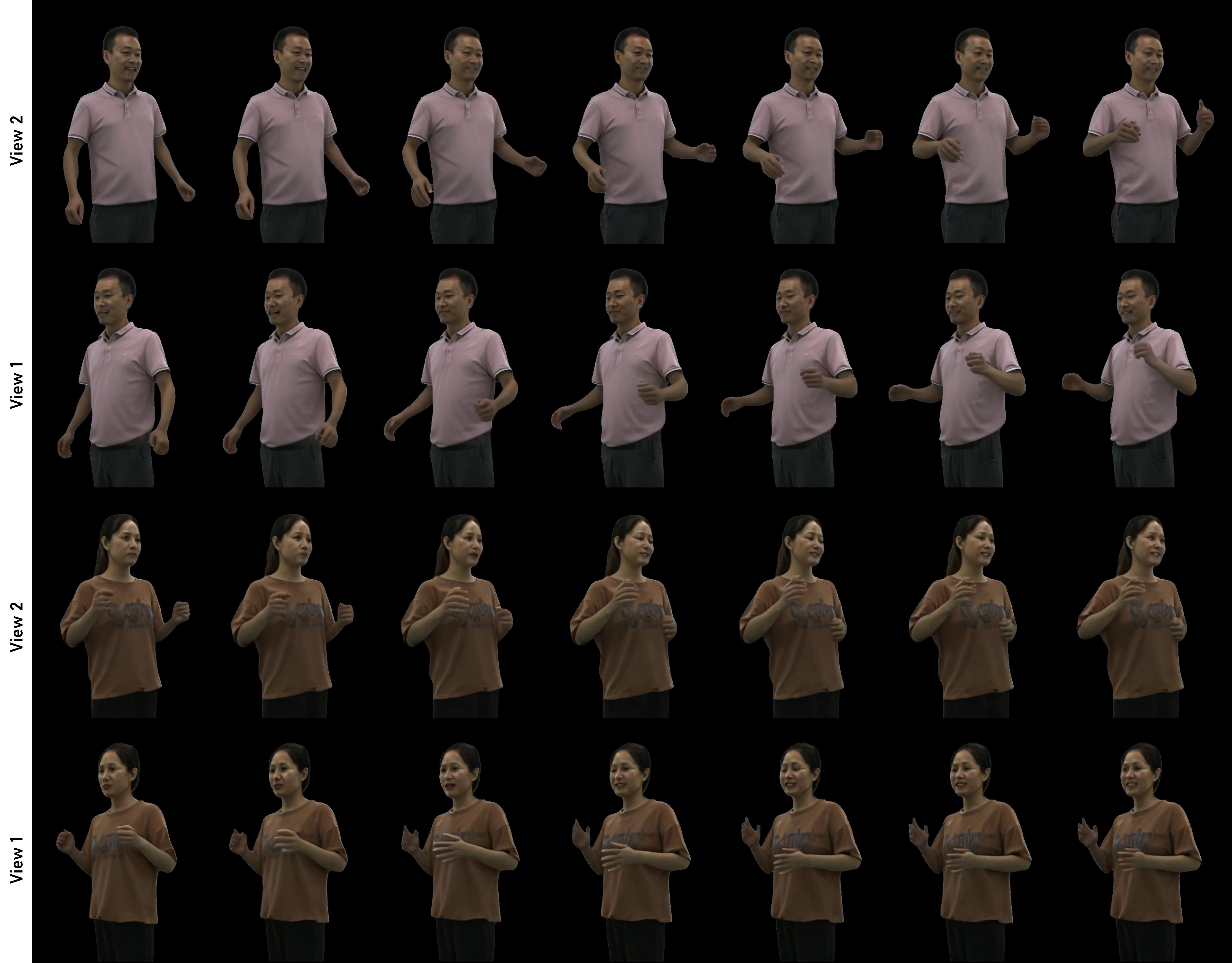}
    \caption{Our method is capable of synthesizing vivid speaking videos with high temporal stability and view consistency.}
    \label{fig:view_consis_talking}
\end{figure*}
Throughout the training phase, the mask encoder, denoted as $\mathcal{E}_m$, processes the $i$-th human mask $\textbf{m}_i$, while the concurrent skeleton encoder, $\mathcal{E}_p$, handles both the $i$-th and $j$-th skeletons $\textbf{p}_i$ and $\textbf{p}_j$. The resultant triplet of the encoded representations is then fed into the decoder, $\mathcal{D}$, which leverages this fused information with the objective of accurately predicting the $j$-th human mask $\textbf{m}_j$:
\begin{equation}
    \bm{{\hat{m}}_j} = \mathcal{D}(\mathcal{E}_m(\bm{m}_i), \mathcal{E}_p(\bm{p}_i), \mathcal{E}_p(\bm{p}_j))
\end{equation}
\begin{equation}
    \mathcal{L}_m =  \mathbbm{E}_i(\mathcal{L}_{kl}(\bm{m}_i, \mathcal{E}_m, \mathcal{E}_p) + \mathcal{L}_{rec}(\bm{m}_i, \bm{{\hat{m}}_i}, \mathcal{D})) 
\end{equation}
where $\mathcal{L}_{rec}$ represents mean square error loss and $\mathcal{L}_{kl}$ represents Kullback-Leibler loss.
Through this strategic partitioning and specialized attention, our model achieves a heightened capability to retain and separately process diverse semantic elements within an image.

\subsection{Human Video Rendering}
\label{Method_render}

Similar to \cite{corona2024vlogger, zhu2024champ}, we extract motion guidance from 3D human parametric models and integrate it into a video diffusion model for subsequent rendering. Specifically, the rendered pose sequences are processed by a lightweight pose guider, composed of four convolutional layers, to incorporate motion guidance. The resulting pose guidance is then masked by an image region mask and added to the noise latent, which is subsequently fed into the corresponding mask-guided experts within the denoising UNet.
For appearance guidance, the reference image is encoded using both a VAE encoder and a CLIP image encoder, with the output subsequently fed into the ReferenceNet. The ReferenceNet shares the same architecture as the denoising UNet, except that it omits the temporal layer. To enhance 3D consistency, we compute the distance embedding between the current input view and each view-guided expert, which is then used as additional input.
The training of pose-guided diffusion render network \cite{hu2023animate} can be formulated as follows:
\begin{equation}
    \mathcal{L}_x = \mathbbm{E}_{\epsilon, \textbf{x}_t, \textbf{p}, \textbf{m}, \textbf{v}, t} (|| \bm{\epsilon} - \bm{\epsilon}_{\theta}(\textbf{x}_{t}, \textbf{p}, \textbf{m}, \textbf{v}, t)||)
\end{equation}
where $\textbf{x}_{t}$ is the noisy human images at timestep \textit{t} and \textbf{v} indicate the view embeddings. 
By enhancing the differentiation among image regions and viewpoints, our framework fosters enhanced rendering consistency and 3D consistency, ultimately yielding improved visual outputs and overall stability.

\begin{table}[!htbp]
\caption{Statistics of existing audio-visual human datasets.}
\begin{adjustbox}{width=0.47\textwidth}

\begin{tabular}{c|c|c|c|c|c}
\bottomrule

Dataset Name   & Subjects& Length &  Dataset Type &  Cleaned & View \\ 
\hline

Speech2Gesture \cite{ginosar2019learning} & 10 & 144h  & Talking & \XSolid & 1\\
PATS \cite{ahuja2020style} & 25 & 251h  & Talking & \XSolid & 1\\
SHOW \cite{yi2023generating} & 4  & 27h & Talking  & \Checkmark & 1\\
BEATX \cite{liu2023emage}& 30 & 76h  & Talking & \Checkmark & 1\\
Tiktok \cite{jafarian2021learning} & 350  & 1h & Dancing & \Checkmark &1\\
Fashion \cite{zablotskaia2019dwnet} & 600 & 2h  & Dancing & \Checkmark &1\\
CCV2 \cite{porgali2023casual} & 5567 & 440h  & Talking & \XSolid&1\\
MVHumanNet \cite{xiong2024mvhumannet} & 4500 & \~{}1000h & Dancing & \Checkmark &48\\
\hline

Ours & 2203 & 15h & Dancing\&Talking & \Checkmark & 1\~{}12\\

\bottomrule
\end{tabular}
\end{adjustbox}
\label{tab:dataset}
\end{table}

\subsection{Dataset}
\label{Method_dataset}

Recent advancements in diffusion models have greatly facilitated the progress of one-shot human video synthesis \cite{hu2023animate, xu2023magicanimate}. Nonetheless, these models typically necessitate substantial training data and heavily rely on large-scale human video datasets comprising a vast number of distinct identities to guarantee their stability and generalizability. 
However, there are no publicly available datasets that meet sufficient conditions (identities, synchronized audio-visual content, multiple viewpoints, etc.) for learning such a robust model, which largely blocks the development of this area.

To fill this gap, we propose a High-definition Audio-Visual dataset (HDAV), which consists of 2,203 identities and the corresponding detail annotation. 
Specifically, we first collect 3,425 online videos from internet and 1,325 videos from existing datasets.
We adopt a cut detection pipeline \footnote{https://github.com/Breakthrough/PySceneDetect} to detect video jumps and unify the frame rate based on moviepy \footnote{https://github.com/Zulko/moviepy}.
A skeleton detector \cite{yang2023effective} is further adopted to crop the human region.
Next, we manually annotate the detailed properties of each video such as gender and clothes type, and remove invalid videos, resulting in 2,173 identity videos with diverse motion ranges and high clarity.
To facilitate 3d consistency and fill the absence of multi-view talking data, we further record 360 multi-view talking videos corresponding to 30 identities and 12 different views.
The comparisons with existing datasets are shown in Table \ref{tab:dataset}. 
The raw video link, the processed motion sequences, and the detailed property labels will be released for research purposes.

%% file: sections/4_experiments.tex
\begin{figure*}
    \centering
    \includegraphics[width = 1\textwidth]{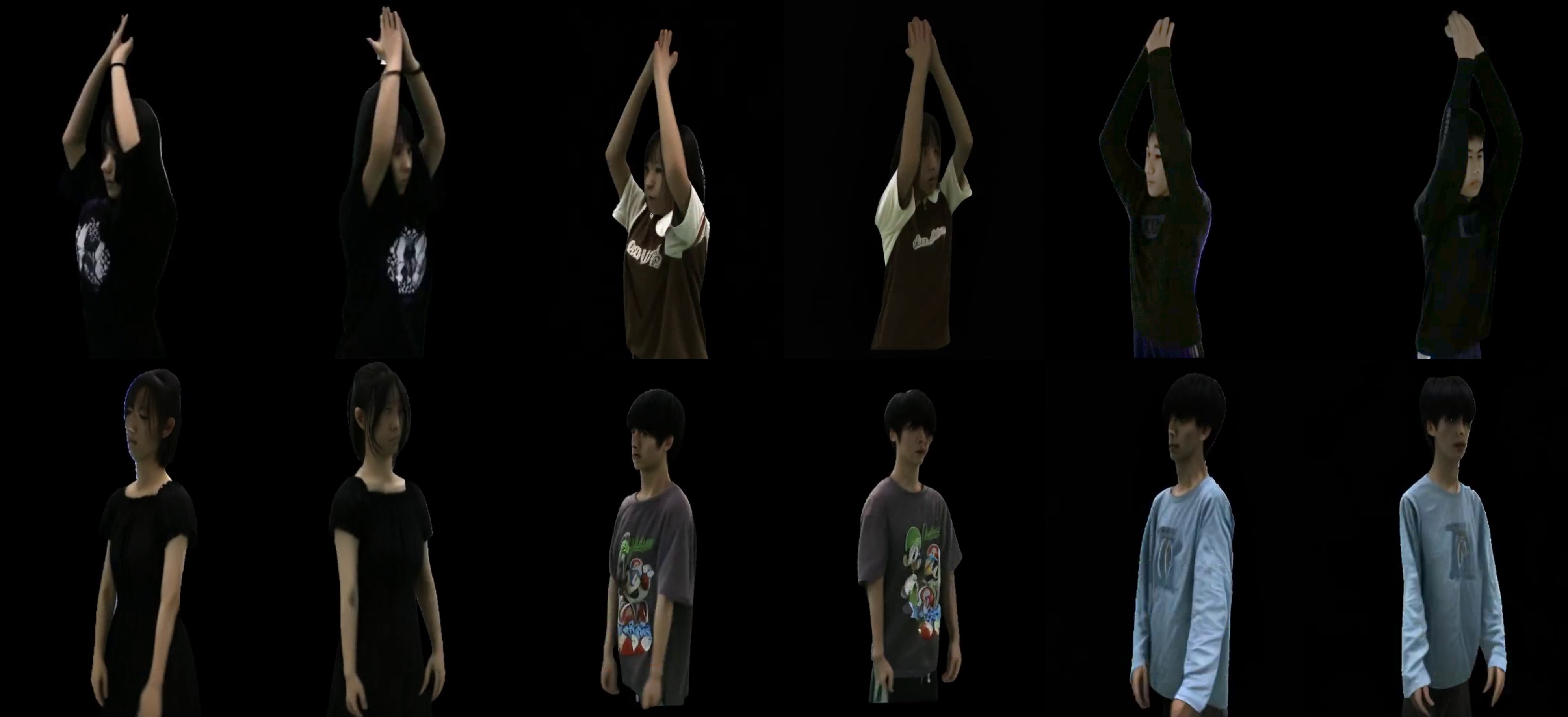}
    \caption{More visualization results of view consistency.}
    \label{fig:view_consis}
\end{figure*}
\section{EXPERIMENTS}
In this section, we give a comprehensive evaluation of our method of fusing the provided audio with a reference image to produce a convincing, life-like 3D talking video. We begin by introducing the experimental setup for evaluation and proceed to present comparative analyses, illustrating how our technique outperforms existing state-of-the-art approaches. 
Further, we evaluate the key components within our method, elucidating their individual impacts on overall performance. 
Some representative examples of our method's efficacy are showcased in Figures \ref{fig:teaser}, \ref{fig:view_consis_talking}, \ref{fig:view_consis}.





\subsection{Experimental Settings}

\noindent\textbf{Dataset}
We trained our render model on a combination of datasets, including our own collected dataset, the TikTok dataset \cite{jafarian2021learning}, the HDTF dataset \cite{zhang2021flow}, and a portion of the MVHumanNet dataset \cite{xiong2024mvhumannet}. From our collected dataset, we randomly selected 100 videos for the validation set and reserved another 200 videos for the test set. Additionally, we randomly chose 10 videos each from the MVHumanNet, TikTok, and HDTF datasets to serve as test sets for video generation and talking head generation, respectively.
For the audio-driven motion generation model, we train it on the Talkshow dataset \cite{yi2023generating} using the official dataset splits provided.




\noindent\textbf{Baselines}
In order to substantiate the superiority of our proposed method, we make comparisons with the following methods, which can be classified into three categories:
\begin{itemize}
    \item \textbf{audio-driven talking video synthesis methods}, including Vlogger \cite{corona2024vlogger}, which is the most related approach that supports generating 2D talking humans with a single portrait image and corresponding audio input. We also make comparisons with MDD \cite{he2024co}, which generates co-speech gesture videos with a motion-decoupled framework trained on person-specific videos.
    To further evaluate the generated face part quality, we make comparisons with one-shot talking-head synthesis methods including Aniportrait \cite{wei2024aniportrait} and SadTalker \cite{zhang2023sadtalker}.
    Since the source code of Vlogger is not available, we adopt the results from its official website \footnote{https://enriccorona.github.io/vlogger/} for comparison.
    For MDD, Aniportrait, and SadTalker, we make comparisons based on the official codes.
    \item \textbf{audio-driven gesture synthesis methods}, including Diffgesture \cite{zhu2023taming}, which makes an early attempt at adopting diffusion networks for generating co-speech gestures, and TalkShow \cite{yi2023generating}, which presents a speech-to-motion generation framework separately models the face, body, and hands.
    We run Diffgesture and TalkShow depending on the official code.
    
\end{itemize}



\noindent\textbf{Implementation Details}
Our diffusion render backbone integrates the principles of ControlNet \cite{zhang2023adding} alongside a Pose Guider module \cite{hu2023animate}. 
The training process for the diffusion render backbone consists of two phases: 1) We initialize the model by activating the corresponding expert within the view-MoE module exclusively with its associated viewpoint data.
2) Following initialization, we proceed to train the entire network uniformly.
This approach ensures a specialized initialization tailored to specific viewpoints before engaging in comprehensive network training.
The training for rendering network is configured with a batch size of 16, executed over a duration of 3 days on eight A800 GPUs.
The optimizer is Adam with a learning rate set at \(1 \times 10^{-5}\).
As for the mask prediction component, we utilize a VAE model that is an adaptation of the Image VAE presented in \cite{rombach2022high}, initializing it with the pre-trained weights provided by the same source to expedite training.
We incorporate an Audio-Gesture Transformer, as detailed in \cite{zhu2023taming}, to serve as the diffusion backbone for our co-gesture generation network. 
The training for this network is configured with a batch size of 128, executed over a duration of 20 hours on an RTX 3090 GPU. We adopt Adam as the optimizer, with a learning rate set at \(1 \times 10^{-4}\).





\begin{figure*}
    \centering
    \includegraphics[width = 1\textwidth]{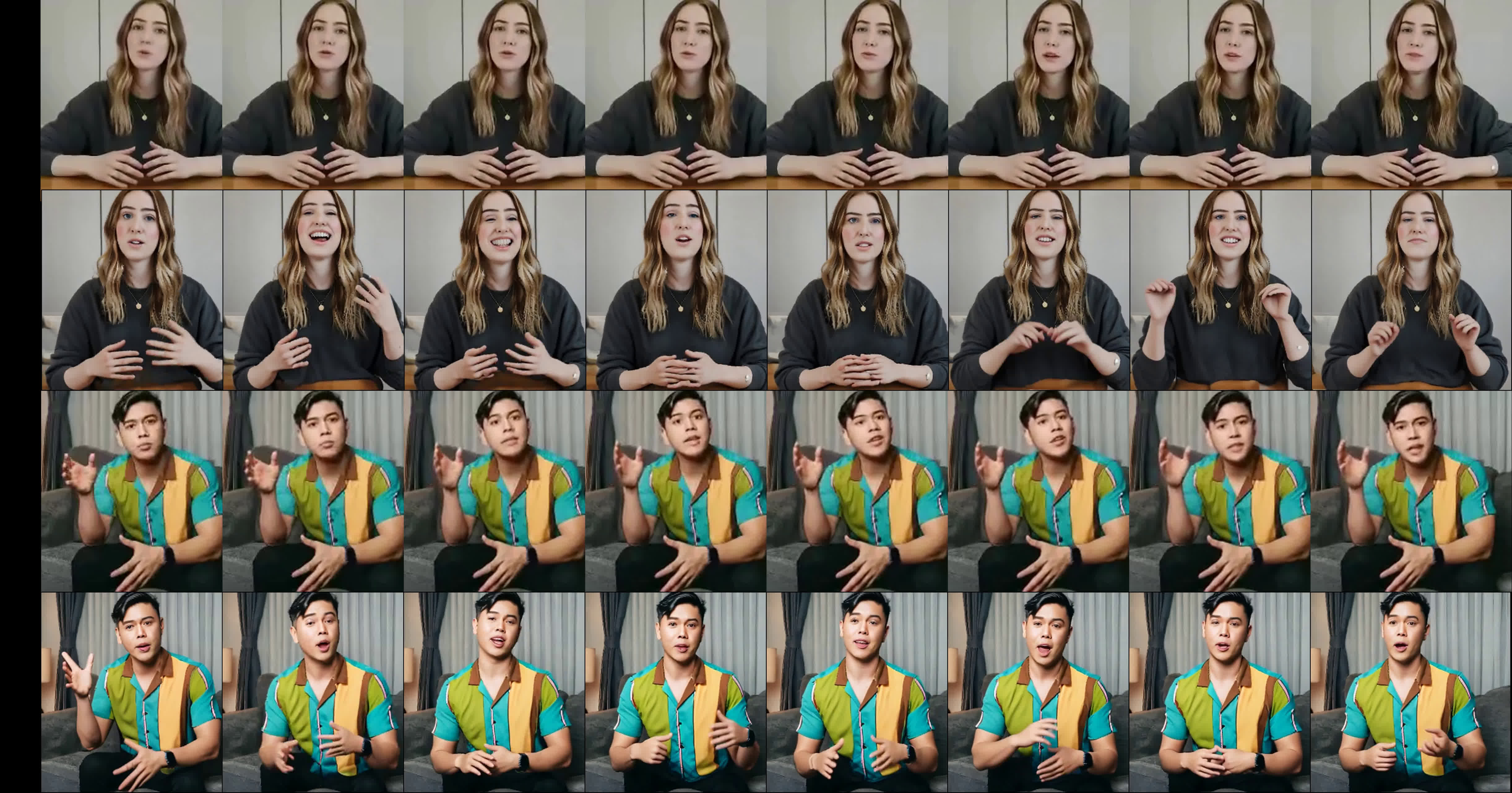}
    \caption{Visualization comparisons with one-shot talking human synthesis method Vlogger \cite{corona2024vlogger}. Our generated outputs exhibit a broader spectrum of body motion diversity, thereby augmenting the overall expressivity and qualitative richness.}
    \label{fig:vlogger}
\end{figure*}
\noindent\textbf{Metrics}
Following previous studies, we adopt three widely used metrics, including \textbf{FID} (Fréchet Inception Distance) \cite{heusel2017gans}, 
\textbf{LPIPS} (Learned Perceptual Image Patch Similarity) \cite{zhang2018unreasonable}, and \textbf{t-LPIPS} \cite{yoon2022learning} as evaluative metrics for visual quality.
We adopt CLIP-D to evaluate the 3D consistency of generated 3D videos. 
As for motion quality, we employ \textbf{FGD} (Fr´echet Gesture Distance) \cite{yoon2020speech}, \textbf{Dir} (Diversity) \cite{li2021audio2gestures}, and \textbf{LVD} (Landmark Velocity Difference) \cite{zhou2020makelttalk}. 
To evaluate the lip synchronization, we adopt \textbf{LSE-C} (Lip Sync Error Confidence) \cite{prajwal2020lip} and \textbf{LSE-D} (Lip Sync Error Distance) \cite{prajwal2020lip} as evaluation metrics.
Given the potential inconsistency between objective metrics and human perception \cite{ijcai2023p650, he2024co}, we adopt subjective metrics including \textbf{Diversity}, \textbf{Synchrony}, \textbf{Clarity}, and \textbf{Overall Quality} in our user study. 
The \textbf{Diversity} and \textbf{Synchrony} metrics emphasize the variety of the generated gestures and the consistency between the audio input and the resultant gestures, respectively. In contrast, the \textbf{Clarity} and \textbf{Overall Quality} metrics concentrate on detailed visual fidelity and the holistic viewing experience.

\begin{table}
\caption{User study comparsion with Vlogger. The ratings for gesture diversity, synchrony between speech and lip motion, detailed visual quality, and overall quality are scored on a scale from 1 to 5, with 5 being the highest. 
The \textbf{bolded} numbers represent the best results.
}
\begin{adjustbox}{width=0.47\textwidth}
\setlength{\tabcolsep}{5pt}
\begin{tabular}{c|cccc}
\bottomrule

Methods & Diversity & Synchrony  & Clarity & Overall Quality  \\ 
\hline
Vlogger \cite{corona2024vlogger}  & 1.88 &  2.2 &  2.44 &  2.68 \\
\hline

Ours  & \textbf{ 4.02 }&  \textbf{3.5} &  \textbf{4.46} &  \textbf{4.1} \\

\bottomrule
\end{tabular}
\end{adjustbox}
\label{tab:Vlogger_user_study}
\end{table}

\begin{table}
\caption{Comparisons with state-of-the-art talking head methods on HDTF dataset. The \textbf{bolded} numbers represent the best results, while the \underline{underlined} numbers indicate the second-best results.
}
\begin{adjustbox}{width=0.47\textwidth}
\setlength{\tabcolsep}{5pt}
\begin{tabular}{c|cc|ccc}
\bottomrule
\multicolumn{1}{c|}{\multirow{2}{*}{\textbf{Methods}}} & \multicolumn{2}{c|}{\textbf{Lip Sync}} & \multicolumn{3}{c}{\textbf{Video Quality}} \\
\cline{2-6}
\multicolumn{1}{c|}{}& \textbf{LSE-C} $\uparrow$ & \textbf{LSE-D} $\downarrow$  & \textbf{FID} $\downarrow$ & \textbf{LPIPS} $\downarrow$  & \textbf{t-LPIPS} $\downarrow$     \\ 
\hline
Ground Truth   & 8.29 & 6.75  & 0 & 0 & 0 \\
SadTalker  \cite{zhang2023sadtalker} &  \textbf{7.57}  &  \textbf{7.54}  &  \underline{22.36}& \underline{0.124} &  0.752  \\
Aniportrait  \cite{wei2024aniportrait}     & 3.51 & 11.11  & 32.86 & 0.225 &\underline{0.641 } \\
\hline
Ours    &\underline{7.17} & \underline{7.93}  & \textbf{17.36} &  \textbf{0.058} & \textbf{0.591}\\
\bottomrule
\end{tabular}
\end{adjustbox}
\label{tab:HDTF}
\end{table}

\begin{figure*}
    \centering
    \includegraphics[width = 1\textwidth]{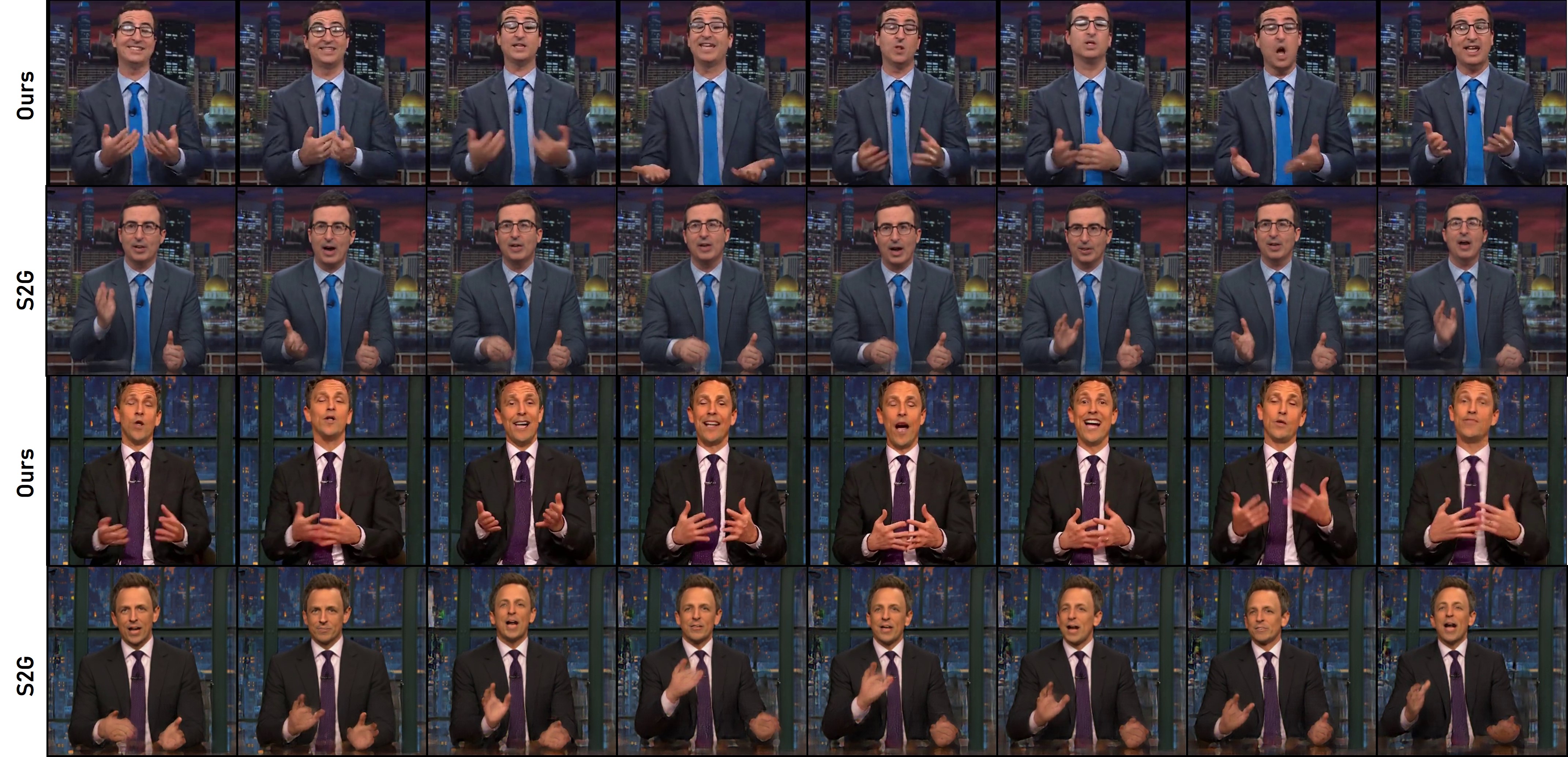}
    \caption{Visualization comparisons with person-specific talking human synthesis method S2G \cite{he2024co}.}
    \label{fig:compare_pats}
\end{figure*}
\begin{figure}
    \centering
    \includegraphics[width = 0.47\textwidth]{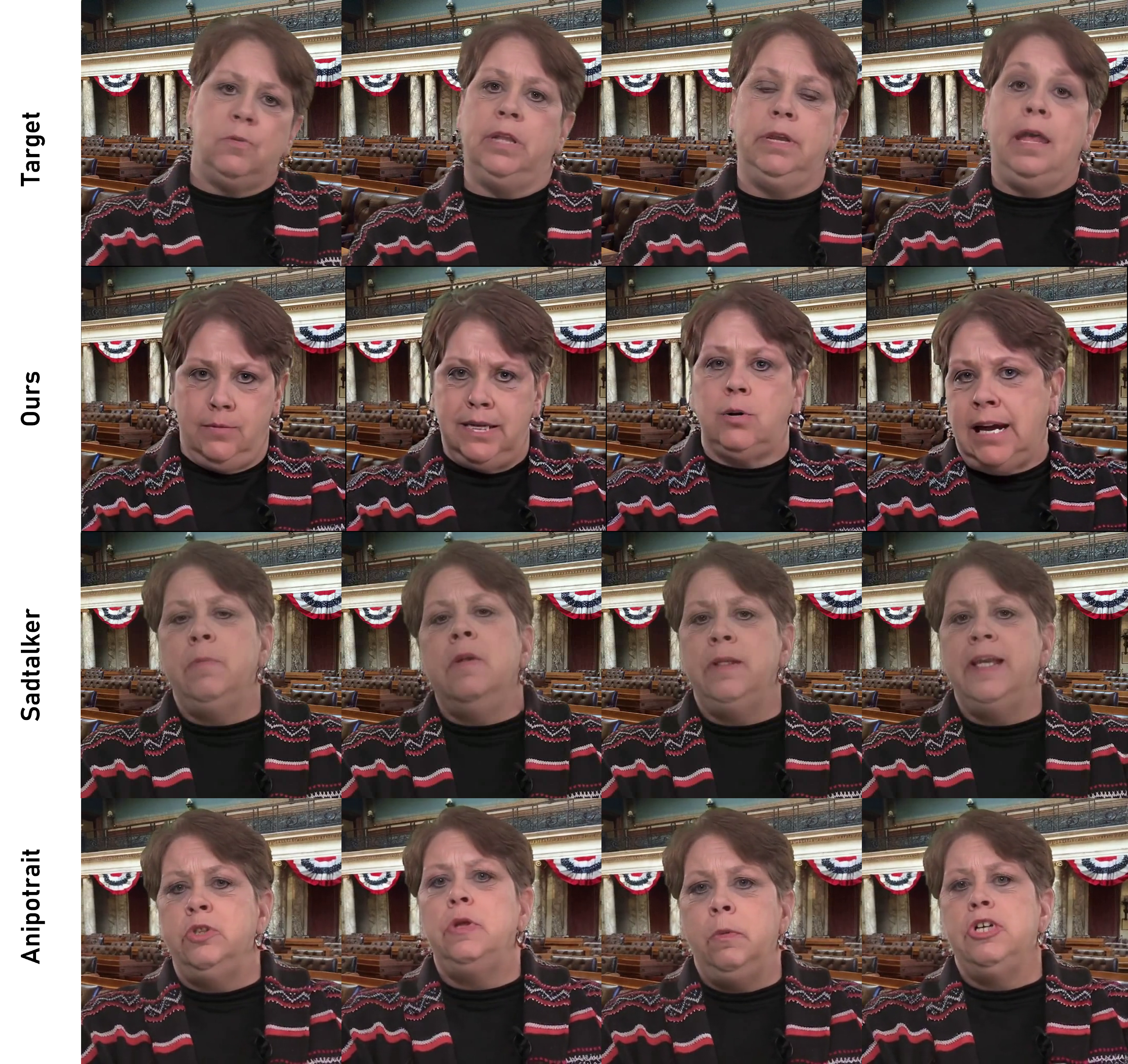}
    \caption{Visualization comparisons with talking head synthesis methods Sadtalker \cite{zhang2023sadtalker} and Aniportrait \cite{wei2024aniportrait}.}
    \label{fig:compare_face}
\end{figure}
\begin{table}
\caption{Comparisons with state-of-the-art person-specific talking human methods on PATS dataset. The \textbf{bolded} numbers represent the best results.}
\begin{adjustbox}{width=0.47\textwidth}
\setlength{\tabcolsep}{5pt}
\begin{tabular}{c|cc|cccc}
\bottomrule
\multicolumn{1}{c|}{\multirow{2}{*}{\textbf{Methods}}} & \multicolumn{2}{c|}{\textbf{Lip Sync}} & \multicolumn{3}{c}{\textbf{Video Quality}} \\
\cline{2-6}
\multicolumn{1}{c|}{}& \textbf{LSE-C} $\uparrow$ & \textbf{LSE-D} $\downarrow$  & \textbf{FID} $\downarrow$ & \textbf{LPIPS} $\downarrow$  & \textbf{t-LPIPS} $\downarrow$     \\ 
\hline
Ground Truth   & 0.572 & 11.99  & 0 & 0 & 0 \\
S2G \cite{he2024co}  &  0.317  &  \textbf{11.65}  & 87.79 & 0.316 &  \textbf{1.118 } \\
\hline
Ours    & \textbf{0.465} & 12.26  & \textbf{64.16} & \textbf{0.315} & 2.491\\
\bottomrule
\end{tabular}
\end{adjustbox}
\label{tab:compare_pats}
\end{table}
\begin{figure}
    \centering
    \includegraphics[width = 0.47\textwidth]{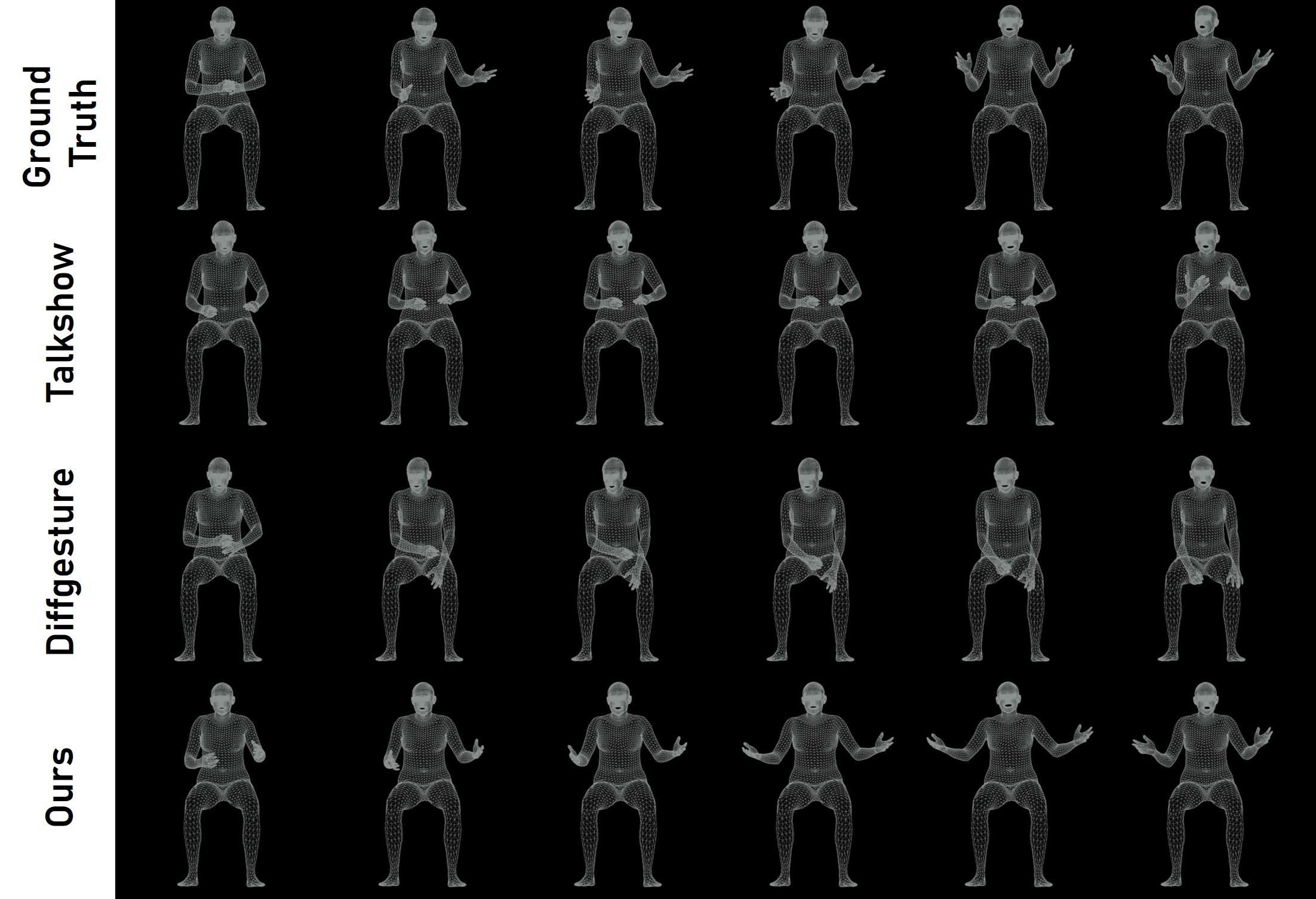}
    \caption{Visualization comparisons with audio-driven motion generation methods Diffgesture \cite{zhu2023taming} and Talkshow \cite{yi2023generating}.}
    \label{fig:audio_gesture}
\end{figure}

\begin{table}
\caption{Comparisons with state-of-the-art audio-driven gesture synthesis methods on Show dataset. The \textbf{bolded} numbers represent the best results, while the \underline{underlined} numbers indicate the second-best results.
}
\begin{adjustbox}{width=0.47\textwidth}
\setlength{\tabcolsep}{5pt}
\begin{tabular}{c|c|c|c}
\bottomrule

Methods & \textbf{LVD} $\downarrow$ & \textbf{L1} $\downarrow$  & \textbf{Dirversity} $\uparrow$  \\ 
\hline
Ground Truth & 0 & 0 & 2.201 \\
SHOW  \cite{yi2023generating}    &   \textbf{0.029} &  \underline{0.072} &  0.356 \\
Diffgesture \cite{zhu2023taming}     & 0.065 & 0.076 &  \underline{1.150} \\

\hline

Ours  &   \underline{0.031} &   \textbf{0.034} &   \textbf{1.721 }\\

\bottomrule
\end{tabular}
\end{adjustbox}
\label{tab:audio_gesture}
\end{table}
\subsection{Results and Comparisons}

\noindent\textbf{Audio-driven Results}
In order to evaluate the motion quality and visual performance of our method, we first compare with Vlogger \cite{corona2024vlogger}, the concurrent work in one-shot 2D talking human synthesis, with illustrative results presented in Figure \ref{fig:vlogger}.
Evidently, Vlogger tend to generate static, invariant poses while our method excels in generating a broad spectrum of dynamic, human-like postures, avoiding the tendency to create inactive and inflexible mean poses, thereby ensuring animations that are both vivid and authentic.
To further evaluate the diversity and quality of our method in comparison with Vlogger, we conducted a user study based on its official website demos since its source code is not available.
The study consists of 30 participants, who grade videos from four aspects on a scale from 1 to 5, with 5 being the best.
As demonstrated in Table \ref{tab:Vlogger_user_study}, our method outperforms Vlogger across all metrics, particularly excelling in motion diversity and detail image quality.

The comparison results between our method and S2G \cite{he2024co} are shown in Figure \ref{fig:compare_pats} and Table \ref{tab:compare_pats}. These results clearly demonstrate that our approach generates more realistic video details, including finer aspects such as finger movements. As shown in Table \ref{tab:compare_pats}, although S2G produces more temporally stable results (lower t-LPIPS) due to its training on target person videos, our method, which uses only a single portrait image as input, achieves competitive or even superior outcomes in most metrics. This further highlights the generalization and high rendering quality of our proposed method.


Further comparisons in Figure \ref{fig:compare_face} and Table \ref{tab:HDTF} reveal that our approach either matches or surpasses the existing one-shot talking head synthesis method across all evaluated metrics. This accomplishment attests to the superior image fidelity and lip synchronization capabilities embedded within our framework.
Although Sadtalker \cite{zhang2023sadtalker} has a better performance in lip synchronization, its scope is inherently limited to facial generation with constrained head movements (see Figure \ref{fig:compare_face} and supplementary video). Conversely, our method not only generates the facial region with high fidelity but also extends to the body region, offering a broader spectrum of body motion diversity, thereby affirming its comprehensive advantage.

\noindent\textbf{Motion Generation Results}
To comprehensively evaluate the motion quality of our method, we conduct comparisons with DiffGesture \cite{zhu2023taming} and Talkshow \cite{yi2023generating}, presenting the results in Figure \ref{fig:audio_gesture} and Table \ref{tab:audio_gesture}. We observe that both our method and Diffgesture generate high-diversity co-speech gestures with largely higher Diversity than Talkshow. However, DiffGesture exhibits inferior performance in L1 and LVD. We attribute the enhanced diversity of co-speech motion generation in our method to the integration of large language models, which exhibit robust generalization capabilities in cross-modal mapping, thereby enriching the diversity and realism of the generated motion sequences.

\subsection{Ablation Studies}

In this part, we conduct an ablation study to evaluate the impact of the key components in our framework.
Specifically, we consider three framework variations, \textit{i.e.}, our framework without view-guided MoE module (w/o MoE-V), our framework without mask MoE-guided module (w/o MoE-M), and our framework without large language model priors (w/o LLM).

\noindent\textbf{Effect of View-guided MoE}
In this part, we conduct an experiment to validate the effectiveness of our view-guided MoE module. 
For each identity in the MVHumanNet test set, we randomly select one view as the reference image and use pose sequences from a different view as the driving signal. This approach allows us to evaluate the model's capability for novel view synthesis.
From Table \ref{tab:ablation_Moe_V}, it is evident that removing the view-guided MoE leads to a significant decrease in all metrics. These observations demonstrate the significant improvement in 3D consistency achieved by the view-guided MoE module, where each expert focuses on a specific viewpoint and collaboratively contributes to the final novel view synthesis quality.


\noindent\textbf{Effect of Mask-guided MoE}
Furthermore, as evidenced in Table \ref{tab:ablation_Moe}, the integration of the mask-guided MoE module alongside the mask prediction module is observed to have a positively beneficial effect on the overall quality of the generated content.
Specifically, without the region segmentation mask involved in the mask-guided MoE module, the performance of the rendering model notably declines, particularly as evaluated by the BRISQUE metric, which directly assesses local image regions and texture quality in the spatial domain. This evidences that our mask-guided MoE module, by incorporating mask information, enhances the distinction among different regions of human images, thereby effectively improving rendering quality and stability.


\noindent\textbf{Effect of LLM prior}
Table \ref{tab:ablation_gesture} demonstrates that the integration of large language model (LLM) priors significantly enhances the quality of motion generation. Specifically, when LLM priors are omitted, the Diversity metrics show a substantial decline, primarily due to the absence of rich, high-level features containing emotional and semantic context. The removal of these priors notably degrades both the diversity of the generated motions and the overall output quality, underscoring their critical importance.


\begin{table}[!htbp]
\caption{Ablation study on View-guided MoE module on MVHumanNet dataset. The \textbf{bolded} numbers represent the best results.}
\begin{adjustbox}{width=0.47\textwidth}

\begin{tabular}{c|c|c|c}
\bottomrule

Methods & \textbf{FID} $\downarrow$ & \textbf{LPIPS} $\downarrow$  & \textbf{t-LPIPS} $\downarrow$  \\ 
\hline

Ours-w/o MoE-V   & 94.05 & 0.496 & 12.93 \\
\hline

Ours  & \textbf{82.08} & \textbf{0.210 }& \textbf{6.03} \\

\bottomrule
\end{tabular}
\end{adjustbox}
\label{tab:ablation_Moe_V}
\end{table}

\begin{table}[!htbp]
\caption{Ablation study on Mask-guided MoE module on tiktok dataset. The \textbf{bolded} numbers represent the best results.}
\begin{adjustbox}{width=0.47\textwidth}

\begin{tabular}{c|c|c|c}
\bottomrule

Methods & \textbf{FID} $\downarrow$ & \textbf{LPIPS} $\downarrow$  & \textbf{t-LPIPS} $\downarrow$  \\ 
\hline

Ours-w/o MoE-M   & 58.47 & 0.319 & 2.023 \\
\hline

Ours  & \textbf{54.22} & \textbf{0.193} & \textbf{1.915} \\

\bottomrule
\end{tabular}
\end{adjustbox}
\label{tab:ablation_Moe}
\end{table}

\begin{table}[!htbp]
\caption{Abaltion study on integration of LLM on our dataset. The \textbf{bolded} numbers represent the best results.}
\begin{adjustbox}{width=0.47\textwidth}

\begin{tabular}{c|c|c|c}
\bottomrule

Methods & \textbf{LVD} $\downarrow$ & \textbf{L1} $\downarrow$  & \textbf{Dirversity} $\uparrow$ \\ 
\hline
Ground Truth & 0 & 0 & 2.201 \\
Our-w/o LLM &  0.033 & 0.039& 1.563\\

\hline

Ours  &  \textbf{0.031} &  \textbf{0.034} & \textbf{ 1.721}\\

\bottomrule
\end{tabular}
\end{adjustbox}
\label{tab:ablation_gesture}
\end{table}

%% file: sections/5_conclusion.tex
\section{Discussion and Conclusion}
\noindent\textbf{Limitations}
The constraints of our methodology can be condensed into two primary points.
First, our model encounters challenges in consistently generating natural and stable hand results, particularly when faced with scenarios involving significant occlusions of hand movements. These extremes in hand gestures can lead to degradation in the quality and realism of the generated output.
Second, since our system relies on a single portrait image to synthesize talking videos, it inherently adheres to a single respect for the individual, thereby failing to establish a comprehensive representation of the identity. 
This limitation may introduce biases into the rendering process.
Consequently, the result videos might occasionally exhibit inconsistencies that could impair the overall viewing experience.

\noindent\textbf{Social Impact}
Our technology empowers the creation of high-quality 3D talking human videos coherent with specify audio and single portrait image. 
However, it could potentially be misused to generate deceptive or misleading content. Thus, it is imperative to proactively consider and implement stringent safeguards against such malicious applications prior to any deployment.

\noindent\textbf{Conclusion}
This paper introduces Stereo-talker, a one-shot 3D talking human synthesis framework. 
To enhance both the temporal and view consistency in rendered videos, we propose integrating prior information through a Mixture of Experts (MoE) module. Specifically, a view-guided MoE steers the rendering model to accurately depict human appearances from diverse viewpoints, while a mask-guided MoE directs the model in effectively distinguishing between different parts of the image.
The mask MoE-guided is complemented by an additional variational autoencoder (VAE) network, trained specifically to predict detailed human masks from skeletal data, which not only improves the accuracy and stability of guided masks at training time but also enables mask guidance at inference time.
By leveraging the power of large language model priors in guiding cross-modal translation, we further enhance the diversity and semantic alignment of co-gesture generation.
Acknowledging the hurdles in training human video generation systems, we also unveil a comprehensive human video dataset comprising 2,203 unique identities, each meticulously annotated with motion sequences and detailed properties. This dataset aims to reduce barriers and facilitate advancements in 3D human generation field.
We believe that our Stereo-talker framework and the accompanying dataset will have profound implications across a spectrum of downstream applications, notably in augmented and virtual reality domains, by enabling more immersive and natural interactions.
